\documentclass[11pt,a4paper]{article}
\usepackage[hyperref]{AACL-IJCNLP2020}
\usepackage{times}
\usepackage{latexsym}
\usepackage{mathtools}
\usepackage{amssymb}
\usepackage{algorithm,algorithmicx}
\usepackage[noend]{algpseudocode}
\usepackage[font=small,labelfont=bf]{caption}
\usepackage[]{todonotes}
\usepackage[nameinlink,capitalize]{cleveref}
\usepackage[inline]{enumitem}
\usepackage{caption}
\usepackage{subcaption}
\usepackage{times}
\usepackage{tikz}

\usepackage{microtype}

\aclfinalcopy 

\title{SimulMT to SimulST: Adapting Simultaneous Text Translation to End-to-End Simultaneous Speech Translation}

\usepackage{amsmath,amsfonts,bm}

\def\eqref#1{equation~\ref{#1}}

\def\1{\bm{1}}

\def\vx{{\bm{x}}}

\def\mA{{\bm{A}}}

\def\mD{{\bm{D}}}

\def\mH{{\bm{H}}}

\def\mX{{\bm{X}}}
\def\mY{{\bm{Y}}}

\DeclareMathAlphabet{\mathsfit}{\encodingdefault}{\sfdefault}{m}{sl}
\SetMathAlphabet{\mathsfit}{bold}{\encodingdefault}{\sfdefault}{bx}{n}

\author{Xutai Ma \\
  Johns Hopkins University  \\
  \texttt{xutai\_ma@jhu.edu} \\
  \And
  Juan Pino \\
  Facebook AI \\
  \texttt{juancarabina@fb.com}
  \And
  Philipp Koehn\\
  Johns Hopkins University\\
  \texttt{phi@jhu.edu}
}

\date{}

\begin{document}
\maketitle
\begin{abstract}
Simultaneous text translation and end-to-end speech translation have recently made great progress but little work has combined these tasks together.
We investigate how to adapt simultaneous text translation methods such as wait-$k$ and monotonic multihead attention to end-to-end simultaneous speech translation by introducing a pre-decision module. A detailed analysis is provided on the latency-quality trade-offs of combining fixed and flexible pre-decision with fixed and flexible policies. We also design a novel computation-aware latency metric, adapted from Average Lagging.
\footnote{The code is available at \url{https://github.com/pytorch/fairseq}}
\end{abstract}
\section{Introduction}


Simultaneous speech translation (SimulST) generates a translation from an input speech utterance before the end of the utterance has been heard. SimulST systems aim at generating translations with maximum quality and minimum latency,
targeting applications such as video caption translations and real-time language interpretation.
While great progress has recently been achieved on both end-to-end speech translation~\cite{ansari-etal-2020-findings} and simultaneous text
translation (SimulMT)~\cite{grissom2014don, gu2017learning, luo2017learning, lawson2018learning, alinejad2018prediction, zheng-etal-2019-simultaneous, zheng-etal-2019-simpler,ma2020monotonic,arivazhagan-etal-2019-monotonic, arivazhagan-etal-2020-translation}, little work has combined the two tasks together~\cite{ren-etal-2020-simulspeech}.

End-to-end SimulST models feature a smaller model size, greater inference speed and fewer compounding errors compared to their cascade counterpart, which perform streaming speech recognition followed by simultaneous machine translation. In addition, it has been demonstrated that end-to-end SimulST systems can have lower latency than cascade systems~\cite{ren-etal-2020-simulspeech}.

In this paper, we study how to adapt methods developed for SimulMT to end-to-end SimulST.
To this end, we introduce the concept of pre-decision module. Such module guides how to group encoder states into meaningful units prior to making a READ/WRITE decision. A detailed analysis of the latency-quality trade-offs when combining a fixed or flexible pre-decision module with a fixed or flexible policy is provided. We also introduce a novel computation-aware latency metric, adapted from Average Lagging (AL)~\cite{ma-etal-2019-stacl}.

\section{Task formalization}
\label{sec:task}
A SimulST model takes as input a sequence of acoustic features $\mX=[\vx_1,...\vx_{|\mX|}]$ extracted from speech samples every $T_{s}$ ms,
and generates a sequence of text tokens $\mY=[y_1,...,y_{|\mY|}]$ in a target language.
Additionally, it is able to generate $y_i$ with only partial input
$\mX_{1:n(y_i)}=[\vx_1,...\vx_{n(y_i)}]$,
where $n(y_i) \leq |\mX|$ is the number of frames needed to generate the $i$-th target token $y_i$.
Note that $n$ is a monotonic function, i.e.\  $n(y_{i-1}) \leq n(y_i)$.

A SimulST model is evaluated with respect to quality, using BLEU~\cite{papineni2002bleu}, and latency. We introduce two latency evaluation methods for SimulST that are adapted from SimulMT.
We first define two types of delays to generate the word $y_i$, a
computation-aware (CA) and a non computation-aware (NCA) delay.
The CA delay of $y_i$, $d_{\text{CA}}(y_i)$,
is defined as the time that elapses (speech duration) from the beginning of the process to the prediction of $y_i$,
while the NCA delay for $y_i$ $d_{\text{CA}}(y_i)$ is defined by
$d_{\text{NCA}}(y_i) = T_{\text{s}} \cdot n(y_i)$.
Note that $d_\text{NCA}$ is an ideal case for $d_\text{CA}$ where the computational time for the model is ignored.
Both delays are measured in milliseconds.
%
Two types of latency measurement, $L_{CA}$ and $L_{NCA}$, are calculated accordingly: $L = \mathcal{C}(\mD)$ where $\mathcal{C}$ is a latency metric and $\mathbf{D}=[d(y_1), ..., d(y_{|\mY|})]$.

To better evaluate the latency for SimulST, we introduce a modification to AL. We assume an oracle system that can perform perfect simultaneous translation
for both latency and quality, while in \citet{ma-etal-2019-stacl} the oracle is ideal only from the latency perspective. We evaluate the lagging based on time rather than steps. The modified AL metric is defined in \cref{eq:al}:
\begin{equation}
    \text{AL} = \frac{1}{\tau(|\mX|)} \sum^{\tau(|\mX|)}_{i=1} d(y_i) - \frac{|\mX|}{|\mY^*|}  \cdot T_s \cdot (i - 1)
    \label{eq:al}
\end{equation}
%
where $|\mY^*|$ is the length of the reference translation, $\tau(|\mX|)$ is the index of the first target token generated when the model has read the full input. There are two benefits from this modification. The first is that latency is measured using time instead of steps, which makes it agnostic to preprocessing and segmentation. The second is that it is more robust and can prevent an extremely low and trivial value when the prediction is significantly shorter than the reference.

\section{Method}
\subsection{Model Architecture}
End-to-end ST models directly map a source speech utterance into a sequence of target tokens.
We use the S-Transformer architecture proposed by \citep{di-gangi-etal-2019-enhancing}, which achieves competitive performance on the MuST-C dataset~\cite{di-gangi-etal-2019-must}.
In the encoder, a two-dimensional attention is applied after the CNN layers and a distance penalty is introduced to bias the attention towards short-range dependencies.

We investigate two types of simultaneous translation mechanisms, flexible and fixed policy. In particular, we investigate monotonic multihead attention~\cite{ma2020monotonic}, which is an instance of flexible policy and the prefix-to-prefix model~\cite{ma-etal-2019-stacl}, an instance of fixed policy, designated by wait-$k$ from now on.

\begin{description}[style=unboxed,leftmargin=0cm]
\item[Monotonic Multihead Attention](MMA)~\cite{ma2020monotonic} extends monotonic attention
\cite{raffel2017online, arivazhagan-etal-2019-monotonic} to Transformer-based models.
Each head in each layer has an independent step probability
$p_{ij}$ for the $i$th target and $j$th source step, and then uses a closed form expected attention for training. 
A weighted average and variance loss were proposed to control the behavior of the attention heads and thus the trade-offs between quality and latency.

\item[Wait-$k$]~\citep{ma-etal-2019-stacl} is a fixed policy that waits for $k$ source tokens,
and then reads and writes alternatively.
Wait-$k$ can be a special case of Monotonic Infinite-Lookback Attention (MILk) \cite{arivazhagan-etal-2019-monotonic} or MMA where the step-wise probability $p_{ij} = 0$ if $j - i < k$ else $p_{ij} = 1$.
\end{description}
\begin{figure}[h]
    \centering
    \includegraphics[width=0.4\textwidth]{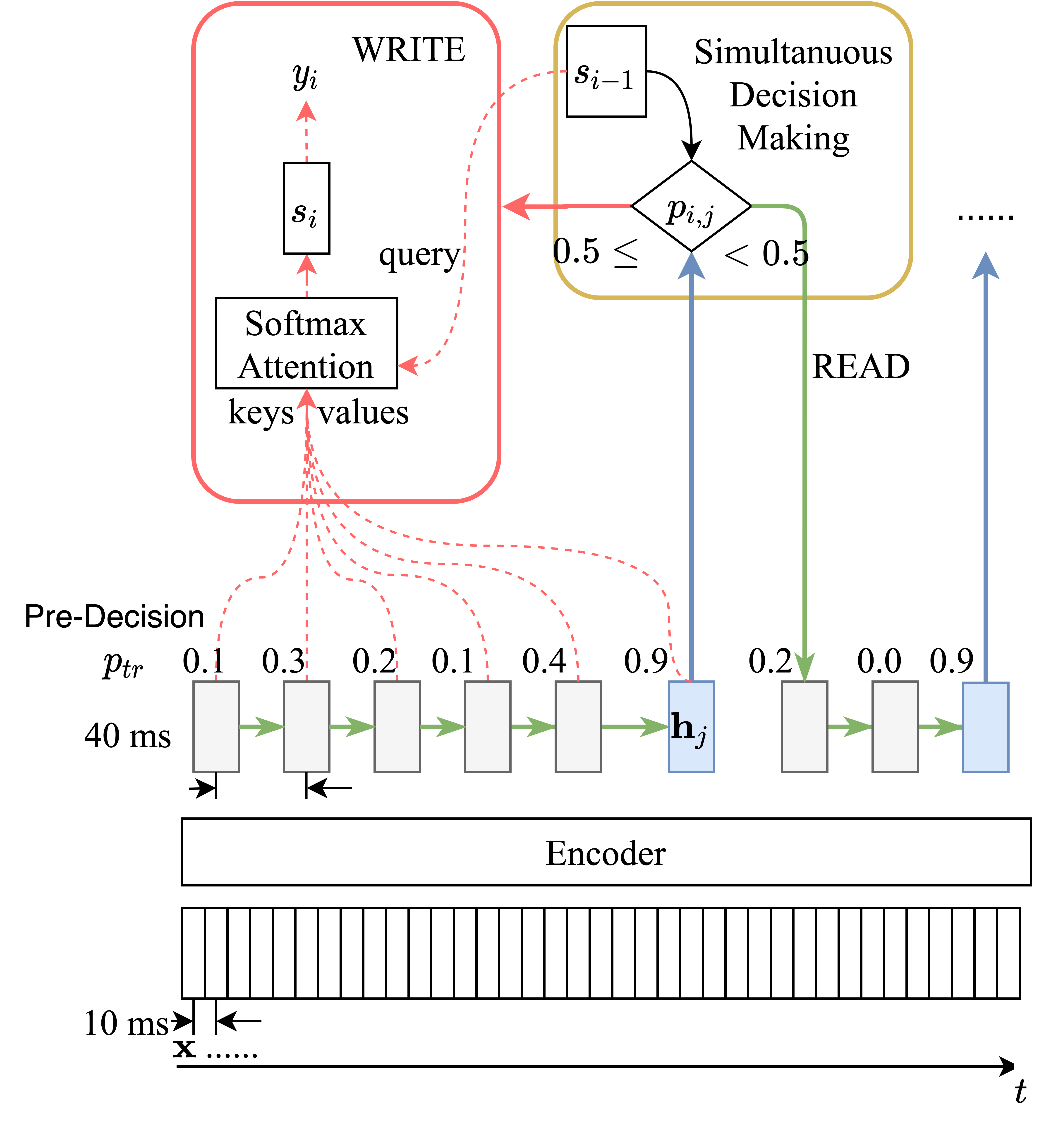}
    \caption{
        Simul-ST architecture with pre-decision module.
        Blue states in the figure indicate when the Simul-ST model triggers the simultaneous decision making process.}
    \label{fig:arch}
\end{figure}
\subsection{Pre-Decision Module}
In SimulMT, READ or WRITE decisions are made at the token (word or BPE) level.
However, with speech input, it is unclear when to make such decisions.
For example, one could choose to read or write after each frame or after
generating each encoder state. Meanwhile, a frame typically only covers 10ms of the input while an encoder state generally covers 40ms of the input (assuming a subsampling factor of 4), while the average length of a word in our dataset is 270ms.
Intuitively, a policy like wait-$k$ will not have enough information
to write a token after reading a frame or generating an encoder state.
In principle, a flexible or model-based policy such as MMA should be able to handle granular input. Our analysis will show, however, that while MMA is more robust to the granularity of the input, it also performs poorly when the input is too fine-grained.

In order to overcome these issues, we introduce the notion of pre-decision module, which groups frames or encoder states, prior to making a decision.
A pre-decision module generates a series of trigger probabilities $p_{tr}$ on each encoder states to
indicate whether a simultaneous decision should be made.
If $p_{tr} > 0.5$, the model triggers the simultaneous decision making,
otherwise keeps reading new frames.
We propose two types of pre-decision module.
\begin{description}[style=unboxed,leftmargin=0cm]
    \item[Fixed Pre-Decision]
 A straightforward policy for a fixed pre-decision module is to trigger simultaneous decision making every fixed number of frames.
 Let $\Delta t$ be the time corresponding to this fixed number of frames, with $\Delta t$ a multiple of $T_s$, and $r_e=\text{int}(|\mX| / |\mH|)$. $p_{tr}$ at encoder step $j$ is defined in \cref{eq:trigger-probablity-fixed}:
    \begin{equation}
        p_{tr}(j) =
        \begin{cases}
            1 & \text{if } \text{mod}(j \cdot r_e \cdot T_s, \Delta t) = 0, \\
            0 & \text{Otherwise.}
        \end{cases}
        \label{eq:trigger-probablity-fixed}
    \end{equation}

\item[Flexible Pre-Decision]
We use an oracle flexible pre-decision
module that uses the source boundaries either at the word or phoneme level. Let $\mA$ be the alignment between encoder states and source labels (word or phoneme). $\mA(h_i)$ represents the token that $h_i$ aligns to.
The trigger probability can then be defined in \cref{eq:trigger-probablity-flexible}:
\begin{equation}
    p_{tr}(j) = \begin{cases}
        0 & \text{if } \mA(h_j) = \mA(h_{j-1}) \\
        1& \text{Otherwise.}
    \end{cases}
    \label{eq:trigger-probablity-flexible}
\end{equation}
\end{description}
\section{Experiments}
We conduct experiments on the English-German portion of the MuST-C dataset~\cite{di-gangi-etal-2019-must}, where source audio, source transcript and target translation are available.
We train on 408 hours of speech and 234k sentences of text data.
We use Kaldi~\cite{povey2011kaldi} to extract 80 dimensional log-mel filter bank features,
computed with a 25$ms$ window size and a 10$ms$ window shift.
For text, we use SentencePiece~\cite{kudo2018sentencepiece} to generate a unigram vocabulary of size 10,000.
We use Gentle\footnote{https://lowerquality.com/gentle/} to generate the alignment between source text and speech
as the label to generate the oracle flexible pre-decision module. Translation quality is evaluated with case-sensitive detokenized BLEU with \textsc{SacreBLEU}~\cite{post-2018-call}.
The latency is evaluated with our proposed modification of AL~\cite{ma-etal-2019-stacl}.
All results are reported on the MuST-C dev set.

All speech translation models are first pre-trained on the ASR task where the target vocabulary is character-based, in order to initialize the encoder.
We follow the same hyperparameter settings from \cite{di-gangi-etal-2019-enhancing}.
We follow the latency regularization method introduced by \cite{ma2020monotonic,  arivazhagan-etal-2019-monotonic}.
The objective function to optimize is
\begin{equation}
    \begin{aligned}
        L = - \text{log} \left( P(\mY| \mX)\right) + \lambda \text{max}\left(\mathcal{C}(\mD), 0\right)
    \end{aligned}
    \label{eq:loss}
\end{equation}
Where $\mathcal{C}$ is a latency metric (AL in this case) and $\mD$ is described in \cref{sec:task}.
Only samples with $\text{AL} > 0$ are regularized to avoid overfitting.
For the models with monotonic multihead attention,
we first train a model without latency with $\lambda_{\text{latency}}=0$.
After the model converges, $\lambda_{\text{latency}}$ is set to a desired value and we continue trining the model until convergence.


\begin{figure*}[t]
    \centering
    \begin{subfigure}[b]{0.45\textwidth}
        \centering
        \includegraphics[width=\textwidth]{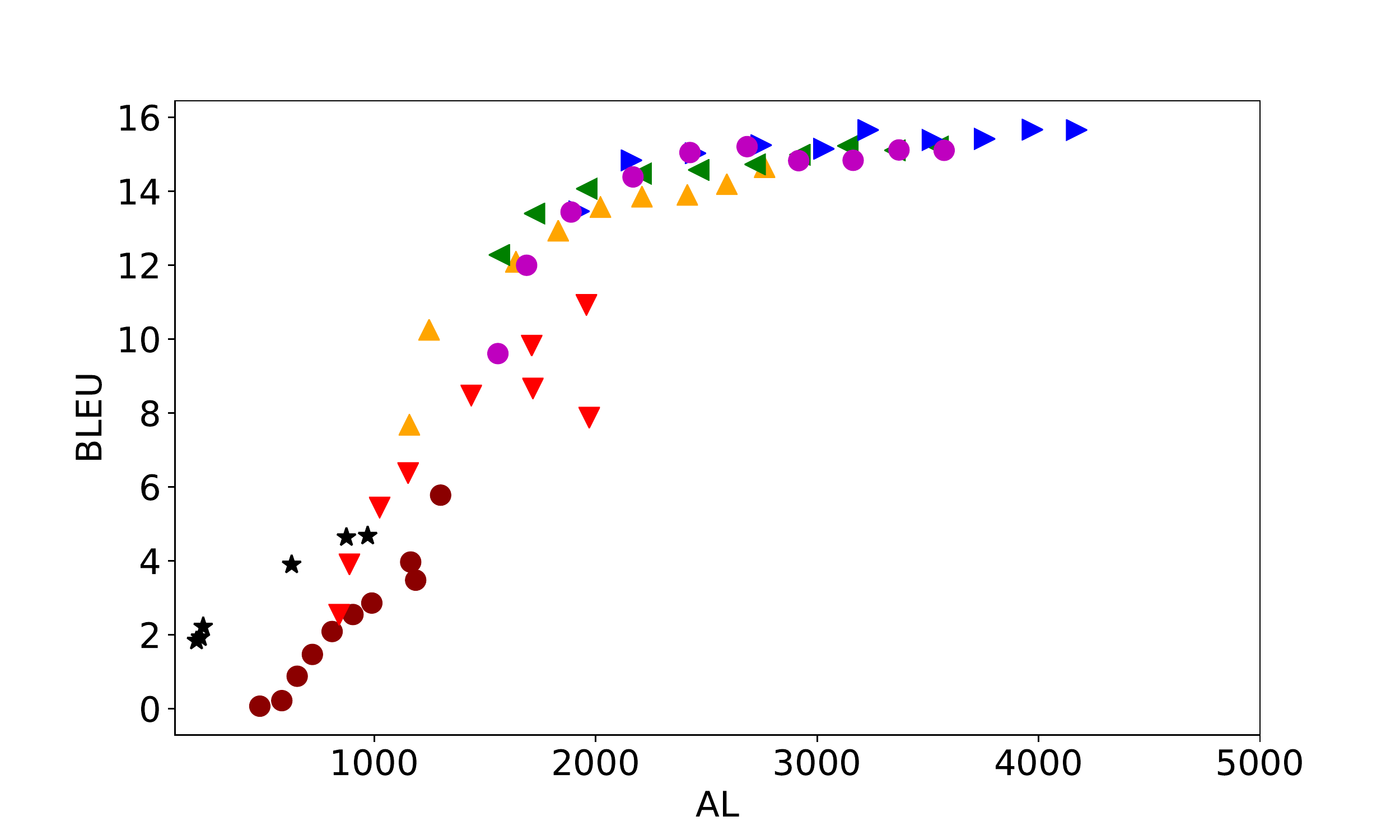}
        \caption{Wait-$k$}
        \label{fig:curve_fixed_policy}
    \end{subfigure}
    \begin{subfigure}[b]{0.45\textwidth}
        \centering
        \includegraphics[width=\textwidth]{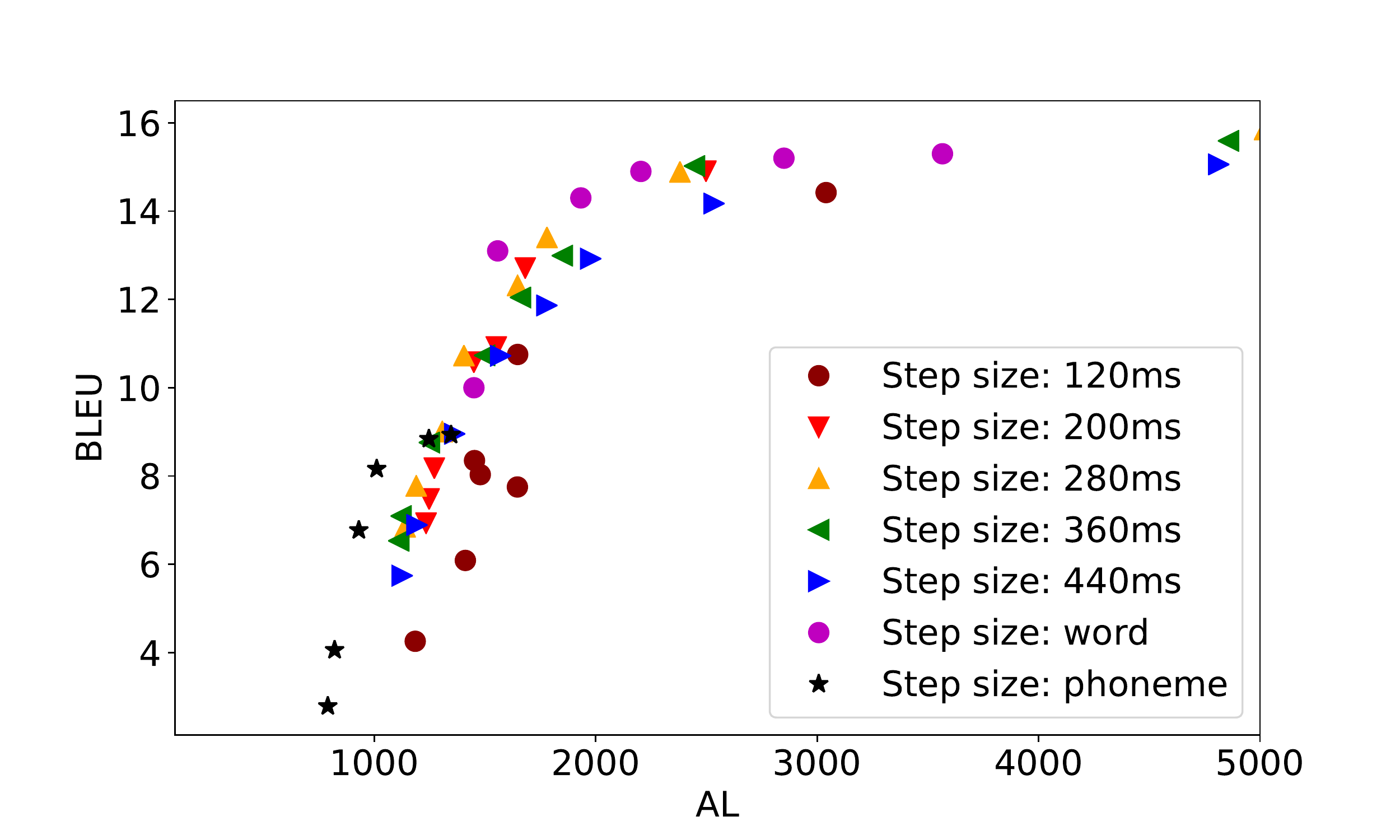}
        \caption{MMA}
        \label{fig:curve_flexible_policy}
    \end{subfigure}
    \caption{Latency-Quality trade-off curves. The unit of AL is millisecond}
    \label{fig:all_curve}
\end{figure*}

\begin{figure}[ht]
    \centering
    \includegraphics[width=0.45\textwidth]{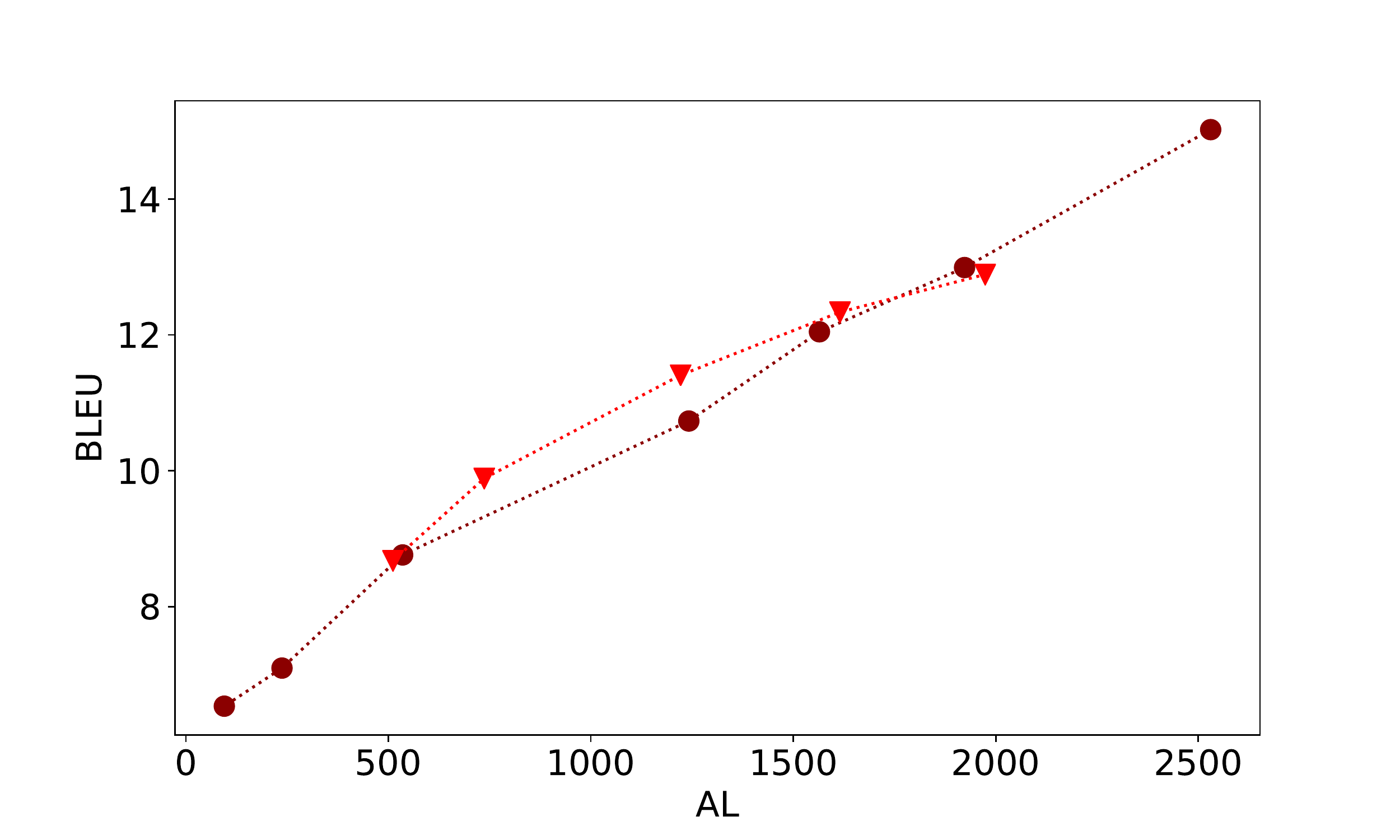}
    \caption{Comparison of best models in four settings}
    \label{fig:best_curve}
\end{figure}

\begin{figure}[ht]
    \centering
    \includegraphics[width=0.45\textwidth]{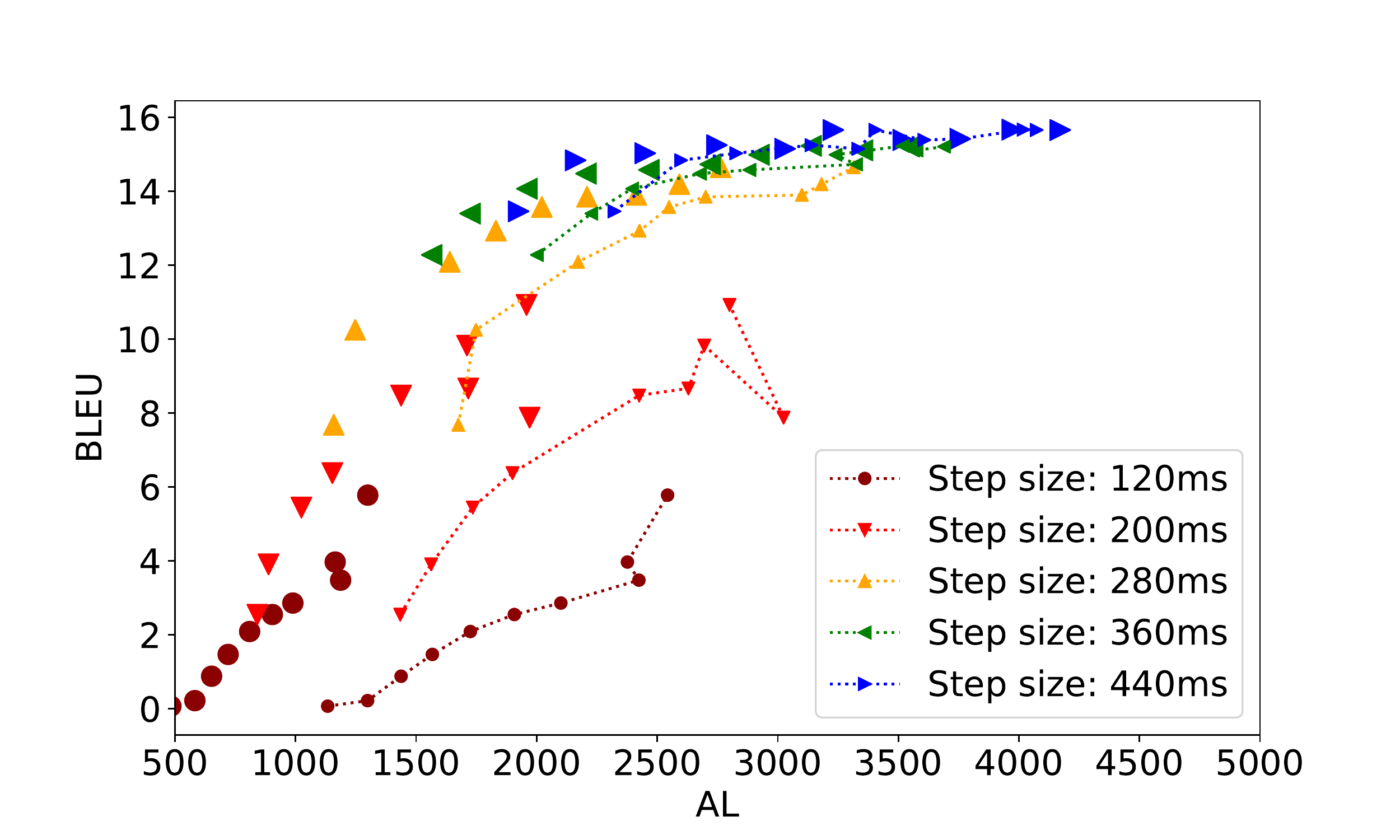}
    \caption{Computation-aware latency for fixed pre-decision + wait-$k$ policy. Points on dotted lines are computation-aware, without lines are non-computation-aware}
    \label{fig:ca_curve}
\end{figure}


The latency-quality trade-offs of the 4 types of model from the combination of fixed or flexible pre-decision with fixed or flexible policy are presented in \cref{fig:all_curve}. The non computation-aware delays are used to calculate the latency metric in order to evaluate those trade-offs from a purely algorithmic perspective.
\begin{description}[style=unboxed,leftmargin=0cm]
\item[Fixed Pre-Decision + Fixed Policy]  \footnote{$k=1,2,3,4,5,6,7,8,9,10$} (\cref{fig:curve_fixed_policy}).
As expected, both quality and latency increase with step size and lagging.
In addition, the latency-quality trade-offs are highly dependent on the step size of the pre-decision module. For example, with step size 120ms, the performance is very poor even with large $k$ because of very limited information being read before writing a target token. Large step sizes improve the quality but introduce a lower bound on the latency.
Note that step size 280ms, which provides an effective latency-quality trade-off compared to other step sizes, also matches the average word length of 271ms. This motivates the study of a flexible pre-decision module based on word boundaries.

\item[Fixed Pre-Decision + Flexible Policy] \footnote{$\lambda=0.001,0.004, 0.01, 0.02, 0.04, 0.06, 0.08, 0.1$} (\cref{fig:curve_flexible_policy})
Similar to wait-$k$, MMA obtains very poor performance with a small step size of 120ms. For other step sizes, MMA obtains similar latency-quality trade-offs, demonstrating some form of robustness to the step size.

\item[Flexible Pre-Decision]
Curve \scalebox{1.5}{$\star$} and \tikz\draw[ violet,fill=violet] (0,0) circle (.5ex);
in figure \cref{fig:all_curve} show latency-quality trade-offs
when the pre-decision module is determined by oracle word or phoneme boundaries.
Note that a SimulST model would not normally have access to this information and that the purpose of this experiment is to guide future design of a flexible pre-decision model.
First, as previously observed, the granularity of the pre-decision greatly influences the latency-quality trade-offs. Models using phoneme boundaries obtain very poor translation quality because those boundaries are too granular, with an average phoneme duration of 77ms.
In addition, comparing MMA and wait-$k$ with phoneme boundaries, MMA is found to be more robust to the granularity of the pre-decision.

\item[Best Curves] The best settings for each approach are compared in \cref{fig:best_curve}. For fixed pre-decision, we choose the setting that has the best quality for each latency bucket of 500ms, while for the flexible pre-decision we use oracle word boundaries.
For both wait-$k$ and MMA, the flexible pre-decision module outperforms the fixed pre-decision module.
This is expected since the flexible pre-decision module uses oracle information in the form of pre-computed word boundaries but provides a direction for future research.
The best latency-quality trade-offs are obtained with MMA and flexible pre-decision from word boundaries.
\end{description}

\subsection{Computation Aware Latency}
\label{sec:cal}
We also consider the computation-aware latency described in \cref{sec:task},
shown in \cref{fig:ca_curve}. The focus is on fixed pre-decision approaches in order to
understand the relation between the granularity of the pre-decision and the computation time.
\cref{fig:ca_curve} shows that as the step size increases, the difference between the NCA and the CA latency shrinks. This is because with larger step sizes, there is less overhead of recomputing the bidirectional encoder states \footnote{This is a common practice in SimulMT where the input length is significantly shorter than in SimulST~\cite{arivazhagan-etal-2019-monotonic, ma-etal-2019-stacl,arivazhagan-etal-2020-translation}}.
We recommend future work on SimulST to make use of CA latency as it reflects a more realistic evaluation, especially in low-latency regimes, and is able to distinguish streaming capable systems.


\section{Conclusion}
We investigated how to adapt SimulMT methods to end-to-end SimulST by introducing the concept of pre-decision module. We also adapted Average Lagging to be computation-aware. The effects of combining a fixed or flexible pre-decision module with a fixed or flexible policy were carefully analyzed. Future work includes building an incremental encoder to reduce the CA latency and design a learnable pre-decision module.


\bibliography{bibliography/simultaneous-translation,bibliography/machine-translation,bibliography/speech-translation,aacl-ijcnlp2020}

\begin{thebibliography}{20}
\expandafter\ifx\csname natexlab\endcsname\relax\def\natexlab#1{#1}\fi

\bibitem[{Alinejad et~al.(2018)Alinejad, Siahbani, and
  Sarkar}]{alinejad2018prediction}
Ashkan Alinejad, Maryam Siahbani, and Anoop Sarkar. 2018.
\newblock Prediction improves simultaneous neural machine translation.
\newblock In \emph{Proceedings of the 2018 Conference on Empirical Methods in
  Natural Language Processing}, pages 3022--3027.

\bibitem[{Ansari et~al.(2020)Ansari, axelrod, Bach, Bojar, Cattoni, Dalvi,
  Durrani, Federico, Federmann, Gu, Huang, Knight, Ma, Nagesh, Negri, Niehues,
  Pino, Salesky, Shi, St{\"u}ker, Turchi, Waibel, and
  Wang}]{ansari-etal-2020-findings}
Ebrahim Ansari, amittai axelrod, Nguyen Bach, Ond{\v{r}}ej Bojar, Roldano
  Cattoni, Fahim Dalvi, Nadir Durrani, Marcello Federico, Christian Federmann,
  Jiatao Gu, Fei Huang, Kevin Knight, Xutai Ma, Ajay Nagesh, Matteo Negri, Jan
  Niehues, Juan Pino, Elizabeth Salesky, Xing Shi, Sebastian St{\"u}ker, Marco
  Turchi, Alexander Waibel, and Changhan Wang. 2020.
\newblock \href {https://www.aclweb.org/anthology/2020.iwslt-1.1} {{FINDINGS}
  {OF} {THE} {IWSLT} 2020 {EVALUATION} {CAMPAIGN}}.
\newblock In \emph{Proceedings of the 17th International Conference on Spoken
  Language Translation}, pages 1--34, Online. Association for Computational
  Linguistics.

\bibitem[{Arivazhagan et~al.(2019)Arivazhagan, Cherry, Macherey, Chiu, Yavuz,
  Pang, Li, and Raffel}]{arivazhagan-etal-2019-monotonic}
Naveen Arivazhagan, Colin Cherry, Wolfgang Macherey, Chung-Cheng Chiu, Semih
  Yavuz, Ruoming Pang, Wei Li, and Colin Raffel. 2019.
\newblock \href {https://www.aclweb.org/anthology/P19-1126} {Monotonic infinite
  lookback attention for simultaneous machine translation}.
\newblock In \emph{Proceedings of the 57th Annual Meeting of the Association
  for Computational Linguistics}, pages 1313--1323, Florence, Italy.
  Association for Computational Linguistics.

\bibitem[{Arivazhagan et~al.(2020)Arivazhagan, Cherry, Macherey, and
  Foster}]{arivazhagan-etal-2020-translation}
Naveen Arivazhagan, Colin Cherry, Wolfgang Macherey, and George Foster. 2020.
\newblock \href {https://www.aclweb.org/anthology/2020.iwslt-1.27}
  {Re-translation versus streaming for simultaneous translation}.
\newblock In \emph{Proceedings of the 17th International Conference on Spoken
  Language Translation}, pages 220--227, Online. Association for Computational
  Linguistics.

\bibitem[{Di~Gangi et~al.(2019{\natexlab{a}})Di~Gangi, Cattoni, Bentivogli,
  Negri, and Turchi}]{di-gangi-etal-2019-must}
Mattia~A. Di~Gangi, Roldano Cattoni, Luisa Bentivogli, Matteo Negri, and Marco
  Turchi. 2019{\natexlab{a}}.
\newblock \href {https://doi.org/10.18653/v1/N19-1202} {{M}u{ST}-{C}: a
  {M}ultilingual {S}peech {T}ranslation {C}orpus}.
\newblock In \emph{Proceedings of the 2019 Conference of the North {A}merican
  Chapter of the Association for Computational Linguistics: Human Language
  Technologies, Volume 1 (Long and Short Papers)}, pages 2012--2017,
  Minneapolis, Minnesota. Association for Computational Linguistics.

\bibitem[{Di~Gangi et~al.(2019{\natexlab{b}})Di~Gangi, Negri, Cattoni, Dessi,
  and Turchi}]{di-gangi-etal-2019-enhancing}
Mattia~Antonino Di~Gangi, Matteo Negri, Roldano Cattoni, Roberto Dessi, and
  Marco Turchi. 2019{\natexlab{b}}.
\newblock \href {https://www.aclweb.org/anthology/W19-6603} {Enhancing
  transformer for end-to-end speech-to-text translation}.
\newblock In \emph{Proceedings of Machine Translation Summit XVII Volume 1:
  Research Track}, pages 21--31, Dublin, Ireland. European Association for
  Machine Translation.

\bibitem[{Grissom~II et~al.(2014)Grissom~II, He, Boyd-Graber, Morgan, and
  Daum{\'e}~III}]{grissom2014don}
Alvin Grissom~II, He~He, Jordan Boyd-Graber, John Morgan, and Hal
  Daum{\'e}~III. 2014.
\newblock Don’t until the final verb wait: Reinforcement learning for
  simultaneous machine translation.
\newblock In \emph{Proceedings of the 2014 Conference on empirical methods in
  natural language processing (EMNLP)}, pages 1342--1352.

\bibitem[{Gu et~al.(2017)Gu, Neubig, Cho, and Li}]{gu2017learning}
Jiatao Gu, Graham Neubig, Kyunghyun Cho, and Victor~OK Li. 2017.
\newblock Learning to translate in real-time with neural machine translation.
\newblock In \emph{15th Conference of the European Chapter of the Association
  for Computational Linguistics, EACL 2017}, pages 1053--1062. Association for
  Computational Linguistics (ACL).

\bibitem[{Kudo and Richardson(2018)}]{kudo2018sentencepiece}
Taku Kudo and John Richardson. 2018.
\newblock Sentencepiece: A simple and language independent subword tokenizer
  and detokenizer for neural text processing.
\newblock In \emph{Proceedings of the 2018 Conference on Empirical Methods in
  Natural Language Processing: System Demonstrations}, pages 66--71.

\bibitem[{Lawson et~al.(2018)Lawson, Chiu, Tucker, Raffel, Swersky, and
  Jaitly}]{lawson2018learning}
Dieterich Lawson, Chung-Cheng Chiu, George Tucker, Colin Raffel, Kevin Swersky,
  and Navdeep Jaitly. 2018.
\newblock Learning hard alignments with variational inference.
\newblock In \emph{2018 IEEE International Conference on Acoustics, Speech and
  Signal Processing (ICASSP)}, pages 5799--5803. IEEE.

\bibitem[{Luo et~al.(2017)Luo, Chiu, Jaitly, and Sutskever}]{luo2017learning}
Yuping Luo, Chung-Cheng Chiu, Navdeep Jaitly, and Ilya Sutskever. 2017.
\newblock Learning online alignments with continuous rewards policy gradient.
\newblock In \emph{2017 IEEE International Conference on Acoustics, Speech and
  Signal Processing (ICASSP)}, pages 2801--2805. IEEE.

\bibitem[{Ma et~al.(2019)Ma, Huang, Xiong, Zheng, Liu, Zheng, Zhang, He, Liu,
  Li, Wu, and Wang}]{ma-etal-2019-stacl}
Mingbo Ma, Liang Huang, Hao Xiong, Renjie Zheng, Kaibo Liu, Baigong Zheng,
  Chuanqiang Zhang, Zhongjun He, Hairong Liu, Xing Li, Hua Wu, and Haifeng
  Wang. 2019.
\newblock \href {https://www.aclweb.org/anthology/P19-1289} {{STACL}:
  Simultaneous translation with implicit anticipation and controllable latency
  using prefix-to-prefix framework}.
\newblock In \emph{Proceedings of the 57th Annual Meeting of the Association
  for Computational Linguistics}, pages 3025--3036, Florence, Italy.
  Association for Computational Linguistics.

\bibitem[{Ma et~al.(2020)Ma, Pino, Cross, Puzon, and Gu}]{ma2020monotonic}
Xutai Ma, Juan Pino, James Cross, Liezl Puzon, and Jiatao Gu. 2020.
\newblock \href {https://openreview.net/forum?id=Hyg96gBKPS} {Monotonic
  multihead attention}.
\newblock In \emph{International Conference on Learning Representations}.

\bibitem[{Papineni et~al.(2002)Papineni, Roukos, Ward, and
  Zhu}]{papineni2002bleu}
Kishore Papineni, Salim Roukos, Todd Ward, and Wei-Jing Zhu. 2002.
\newblock Bleu: a method for automatic evaluation of machine translation.
\newblock In \emph{Proceedings of the 40th annual meeting on association for
  computational linguistics}, pages 311--318. Association for Computational
  Linguistics.

\bibitem[{Post(2018)}]{post-2018-call}
Matt Post. 2018.
\newblock \href {https://www.aclweb.org/anthology/W18-6319} {A call for clarity
  in reporting {BLEU} scores}.
\newblock In \emph{Proceedings of the Third Conference on Machine Translation:
  Research Papers}, pages 186--191, Belgium, Brussels. Association for
  Computational Linguistics.

\bibitem[{Povey et~al.(2011)Povey, Ghoshal, Boulianne, Burget, Glembek, Goel,
  Hannemann, Motlicek, Qian, Schwarz et~al.}]{povey2011kaldi}
Daniel Povey, Arnab Ghoshal, Gilles Boulianne, Lukas Burget, Ondrej Glembek,
  Nagendra Goel, Mirko Hannemann, Petr Motlicek, Yanmin Qian, Petr Schwarz,
  et~al. 2011.
\newblock The kaldi speech recognition toolkit.
\newblock In \emph{IEEE 2011 workshop on automatic speech recognition and
  understanding}, CONF. IEEE Signal Processing Society.

\bibitem[{Raffel et~al.(2017)Raffel, Luong, Liu, Weiss, and
  Eck}]{raffel2017online}
Colin Raffel, Minh-Thang Luong, Peter~J Liu, Ron~J Weiss, and Douglas Eck.
  2017.
\newblock Online and linear-time attention by enforcing monotonic alignments.
\newblock In \emph{Proceedings of the 34th International Conference on Machine
  Learning-Volume 70}, pages 2837--2846. JMLR. org.

\bibitem[{Ren et~al.(2020)Ren, Liu, Tan, Zhang, QIN, Zhao, and
  Liu}]{ren-etal-2020-simulspeech}
Yi~Ren, Jinglin Liu, Xu~Tan, Chen Zhang, Tao QIN, Zhou Zhao, and Tie-Yan Liu.
  2020.
\newblock \href {https://www.aclweb.org/anthology/2020.acl-main.350}
  {{S}imul{S}peech: End-to-end simultaneous speech to text translation}.
\newblock In \emph{Proceedings of the 58th Annual Meeting of the Association
  for Computational Linguistics}, pages 3787--3796, Online. Association for
  Computational Linguistics.

\bibitem[{Zheng et~al.(2019{\natexlab{a}})Zheng, Zheng, Ma, and
  Huang}]{zheng-etal-2019-simpler}
Baigong Zheng, Renjie Zheng, Mingbo Ma, and Liang Huang. 2019{\natexlab{a}}.
\newblock \href {https://doi.org/10.18653/v1/D19-1137} {Simpler and faster
  learning of adaptive policies for simultaneous translation}.
\newblock In \emph{Proceedings of the 2019 Conference on Empirical Methods in
  Natural Language Processing and the 9th International Joint Conference on
  Natural Language Processing (EMNLP-IJCNLP)}, pages 1349--1354, Hong Kong,
  China. Association for Computational Linguistics.

\bibitem[{Zheng et~al.(2019{\natexlab{b}})Zheng, Zheng, Ma, and
  Huang}]{zheng-etal-2019-simultaneous}
Baigong Zheng, Renjie Zheng, Mingbo Ma, and Liang Huang. 2019{\natexlab{b}}.
\newblock \href {https://doi.org/10.18653/v1/P19-1582} {Simultaneous
  translation with flexible policy via restricted imitation learning}.
\newblock In \emph{Proceedings of the 57th Annual Meeting of the Association
  for Computational Linguistics}, pages 5816--5822, Florence, Italy.
  Association for Computational Linguistics.

\end{thebibliography}
\bibliographystyle{acl_natbib}
\appendix
\end{document}